\documentclass[conference]{IEEEtran}
\IEEEoverridecommandlockouts

\usepackage{booktabs}
\usepackage{multirow}
\usepackage{graphicx}
\usepackage{cite}
\usepackage{amsmath,amssymb,amsfonts}
\usepackage{algorithmic}
\usepackage{graphicx}
\usepackage{textcomp}
\usepackage{xcolor}
\usepackage[symbol]{footmisc}

\def\BibTeX{{\rm B\kern-.05em{\sc i\kern-.025em b}\kern-.08em
    T\kern-.1667em\lower.7ex\hbox{E}\kern-.125emX}}
\begin{document}

\title{Boosting Knowledge-based Visual Question Answering with Structured Context Reasoning}

\author{
	\IEEEauthorblockN{Qiyou Liu$^{1}$, Yong Zhang$^{2}$, Jianjie Luo$^{1,*}$, Zhenguo Yang$^{1}$, Yi Yu$^{3}$}
	\IEEEauthorblockA{
		$^{1}$School of Computer Science, Guangdong University of Technology, Guangzhou, China\\
		$^{2}$School of Artificial Intelligence, Shenzhen University, Shenzhen, China\\
		$^{3}$Graduate School of Advanced Science and Engineering, Hiroshima University, Hiroshima, Japan\\
		globarstarliu@gmail.com, yongzhang@szu.edu.cn,
		\{jianjieluo, yzg\}@gdut.edu.cn,
		yiyu@hiroshima-u.ac.jp
	}
}

\maketitle

\begin{abstract}
Knowledge-based Visual Question Answering aims to answer questions about an image by integrating external knowledge with visual and textual information. Recent approaches often rely on in-context learning to prompt Large Language Models (LLMs) with multimodal context in a zero-shot or few-shot manner. However, we observe that directly concatenating heterogeneous visual descriptions and retrieved knowledge into long, unstructured prompts often degrades reasoning performance, due to both excessive irrelevant context and the lack of explicit relational structure. In this paper, we propose an LLM-based \underline{\textbf{S}}tructured \underline{\textbf{Co}}ntext \underline{\textbf{Re}}asoning (\textbf{SCoRe}) framework that infers both explicit and implicit relationships for prediction. SCoRe consists of three stages: Context Acquisition, which generates diverse visual notes and retrieves explicit knowledge via an efficient two-stage multimodal retrieval strategy; Context Selection, which filters relevant visual, explicit, and implicit knowledge using LLM-guided selection; and Context Compression, which performs Relational Logic Distillation (RLD) to transform raw text into explicit entity-relation triplets. These relational triplets serve as a concise and structured prompt for final answer prediction. Extensive experiments on the OK-VQA and A-OKVQA benchmarks demonstrate that SCoRe consistently outperforms state-of-the-art methods.
\end{abstract}

\begin{IEEEkeywords}
Knowledge-based Visual Question Answering, Structured Context Reasoning.
\end{IEEEkeywords}

\vspace{-0.5em}
\section{Introduction}
\vspace{-0.2em}
\label{sec:intro}

\footnotetext[1]{J. Luo is the corresponding author.}

Knowledge-based Visual Question Answering (KB-VQA) answers questions by integrating external knowledge with visual and textual information. Compared with conventional VQA, KB-VQA often involves implicit concepts and multi-step reasoning, making effective organization of heterogeneous information crucial for accurate prediction.

\begin{figure}[t]
	\includegraphics[width=\columnwidth]{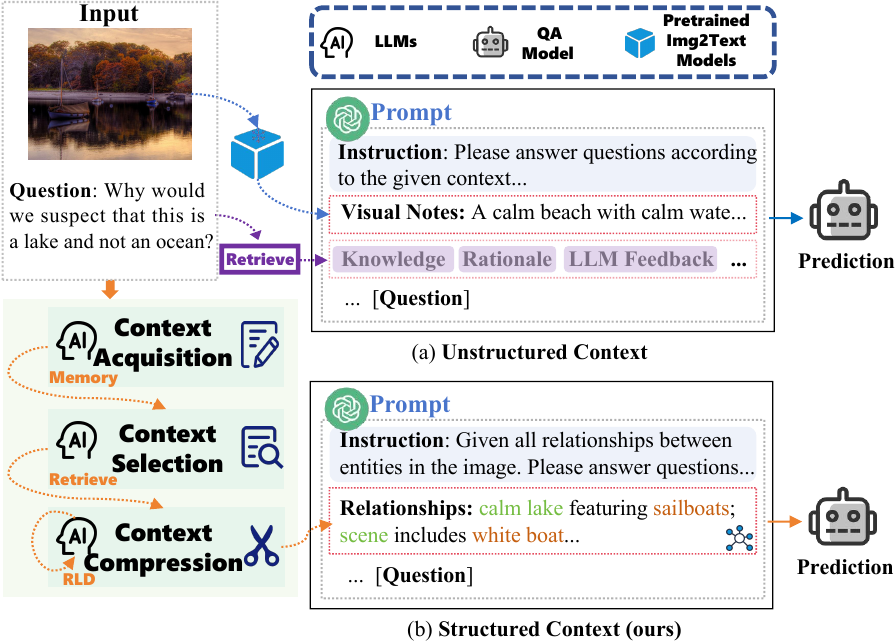}
	\vspace{-2.1em}
	\caption{Two paradigms for LLM-based KB-VQA: (a) Unstructured context reliance, where raw evidence is concatenated into flat, redundant prompts; (b) Our structured context reasoning, which distills multimodal inputs into relational triplets through progressive acquisition, selection, and compression.}
	\label{fig:paradigms}
	\vspace{-1.5em}
\end{figure}

Recent works \cite{yang2022empirical,guo2023images,Chen2024VisualCP, qiang2025enhancing} increasingly leverage Large Language Models (LLMs) to address KB-VQA in zero-shot or few-shot settings by concatenating image descriptions, retrieved knowledge, and task instructions into a single prompt, as illustrated in Fig. \ref{fig:paradigms} (a). Although being effective in some cases, this unstructured prompting paradigm presents inherent limitations. First, the prompt often contains redundant or weakly relevant information, which distracts the model from key evidence. Second, unstructured context fails to explicitly represent relationships among entities, forcing LLMs to implicitly infer complex logical dependencies during reasoning. These issues are particularly pronounced in KB-VQA, where answering questions often depends on understanding relations rather than isolated facts. Several prior methods attempt to alleviate these problems through knowledge selection \cite{cao2024knowledge}, rationale extraction \cite{li2024diversify}, or heuristic filtering \cite{shao2023prompting}. However, such approaches still operate over a flat textual context and do not explicitly model relational structure. 

To this end, we propose Structured Context Reasoning (SCoRe), a framework that explicitly transforms heterogeneous multimodal inputs into compact relational representations tailored for LLM-based reasoning, as illustrated in Fig. \ref{fig:paradigms} (b). Instead of treating context as a flat sequence of text, SCoRe progressively constructs a structured reasoning substrate through three stages. First, the Context Acquisition stage generates diverse visual notes and retrieves explicit external knowledge using an efficient two-stage multimodal retrieval strategy. Second, the Context Selection stage leverages LLMs to identify and retain only information that is relevant to the question, while also eliciting implicit knowledge aligned with the visual content. Finally, the Context Compression stage implements Relational Logic Distillation (RLD) guided through a systematic logical decomposition process, which distills flat text into structured entity-relation triplets, yielding a compact and logic-oriented context for answer prediction. Experiments on OK-VQA and A-OKVQA demonstrate that SCoRe consistently outperforms state-of-the-art zero-shot and few-shot methods. The results indicate that explicitly structuring context into relational representations improves reasoning effectiveness beyond simply increasing or filtering textual context. We show that context structure, rather than context quantity, is the key factor for effective KB-VQA.

In summary, our contributions are as follows:
\begin{itemize}
	\item We propose SCoRe, a structured context reasoning framework for KB-VQA that reformulates heterogeneous multimodal information into compact relational representations, enabling more effective LLM-based reasoning.
	
	\item We design a two-stage progressive multimodal retrieval strategy that significantly reduces the retrieval search space while preserving high recall of relevant knowledge.
	
	\item We propose RLD, a logic-driven inference mechanism that models entity relations to transform evidence into a structured, reasoning-oriented context.
	
	\item We conduct extensive experiments on OK-VQA and A-OKVQA, achieving consistent improvements over strong baselines and validating the effectiveness of structured context representations for KB-VQA.
\end{itemize}

\begin{figure*}[t]
	\centering
	\includegraphics[width=0.96\textwidth]{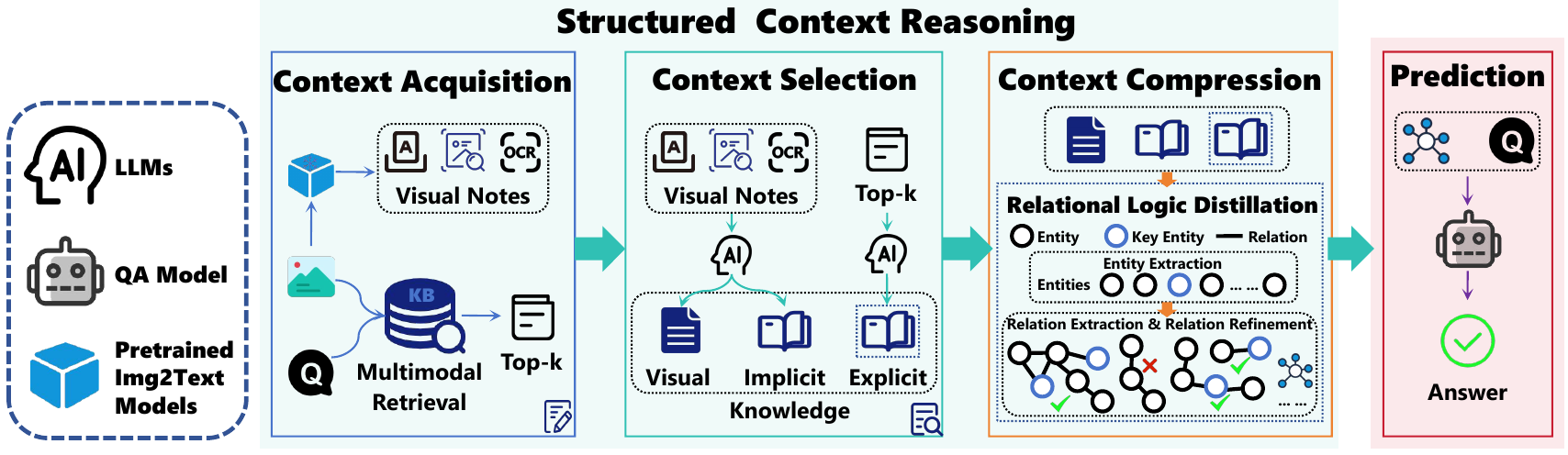}
	\vspace{-0.8em}
	\caption{The proposed LLM-based Structured Context Reasoning (SCoRe) method.}
	\label{fig:framework}
	\vspace{-1.5em}
\end{figure*}

\vspace{-0.5em}
\section{Related Work}
\vspace{-0.2em}
\textbf{Knowledge-based VQA.} Knowledge-based VQA integrates external knowledge with multimodal inputs to answer questions. Traditional methods \cite{lin2022retrieval,wu2022multi} relied on supervised learning with retrieved knowledge (e.g., Wikipedia), but faced limitations in retrieval accuracy and generalization. With the rise of Large Language Models (LLMs) and Large Vision-Language Models (LVLMs), quite a few methods \cite{yang2022empirical,shao2023prompting} have started to use LLMs or LVLMs as reasoning models, leveraging the knowledge they contain to infer answers. The basic method \cite{guo2023images} directly used image descriptions to prompt LLMs. For questions that required comprehensive reasoning, the LLMs may struggle to arrive at answers based solely on image descriptions. The subsequent method KGenVQA \cite{cao2024knowledge} leveraged LLMs to generate relevant knowledge, then combined the image descriptions to answer the question. This may lead to the introduction of a large amount of redundant information. The method DIETCOKE \cite{li2024diversify} leveraged the LLMs to extract rational sentences from the context. This helped LLMs focus more effectively on key information, but improper extraction may lead to the loss of some important details. These methods based on LLMs all employed unstructured context, which resulted in incoherent logic. Moreover, methods such as DIETCOKE \cite{li2024diversify} relied solely on the LLMs' knowledge for question answering, whereas we retrieve explicit knowledge following the PREFLMR \cite{preflmr} and combine it with LLMs' knowledge to predict the answer.

\textbf{Relation Extraction.} Relation Extraction (RE) identifies structured entity dependencies to provide a logical foundation for reasoning. MUMRC \cite{mumrc} recasts multimodal triplet extraction as machine reading comprehension, while TEVLA \cite{tevla} utilizes generative text augmentation for robust vision-language alignment. Recent LLM-based methods like QA4RE \cite{qa4re} elicit zero-shot extraction by aligning RE with common instruction tasks, and OSLLM \cite{osllm} improves consistency via a retrieve-reason-refine framework. Additionally, ERA-Cot \cite{liu2024era} underscores that analyzing entity relationships is vital for enhancing logical depth. Unlike methods focused on standalone extraction accuracy, our SCoRe framework implements Relational Logic Distillation as a systematic logical decomposition process. It distills heterogeneous evidence into structured relational triplets, effectively alleviating cognitive overload and facilitating final answer prediction in KB-VQA.

\section{Methodology}

In this section, we introduce SCoRe in detail. SCoRe includes three stages: Context Acquisition (CA), Context Selection (CS), and Context Compression (CC). Specifically, the CA stage is responsible for retrieving explicit knowledge and converting image information into multi-aspect visual notes. The CS stage retrieves implicit knowledge and extracts the relevant information to obtain a concise context. Subsequently, the CC stage combines such information to extract structured relations. Finally, the QA model produces answers using these relations. The overview of SCoRe is shown in Fig. \ref{fig:framework}.
\subsection{Context Acquisition}
The Context Acquisition stage aims to construct a comprehensive but decomposed evidence pool from both visual inputs and external knowledge sources.

\textbf{Visual Evidence.} To preserve complementary visual cues, we represent the image using three parallel views: object-level attributes, image captions, and OCR-extracted text. Object attributes are extracted using a pre-trained detector, captions are generated via question-guided captioning, and OCR strings are obtained from the image text. Together, these components form a set of visual notes that explicitly separate different types of visual evidence.

\begin{figure}[t]
	\centering
	\includegraphics[width=0.8\columnwidth]{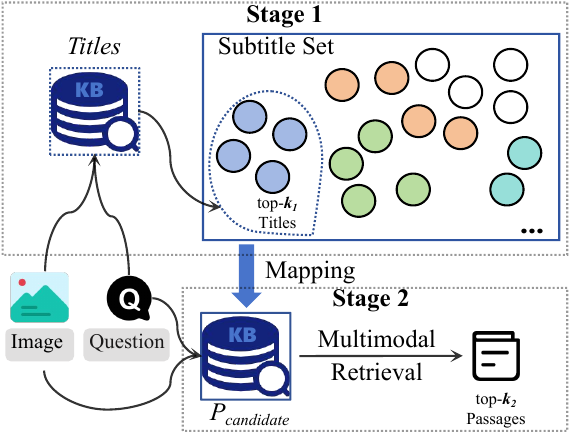}
	\vspace{-0.8em}
	\caption{The two-stage progressive multimodal retrieval strategy.}
	\label{fig:mr}
	\vspace{-1.5em}
\end{figure}

\textbf{Explicit Knowledge Retrieval.} Consistent with previous work, we employ the Google Search corpus and Wikipedia as explicit knowledge bases for OK-VQA and A-OKVQA, respectively. To address the retrieval efficiency bottleneck caused by the massive scale of the Wikipedia, we design a two-stage progressive multimodal retrieval strategy, as illustrated in Fig. \ref{fig:mr}. Specifically, we first utilize a text encoder $F_{L}$ and a visual encoder $F_{V}$ to encode the question $q$ and the image $i$, respectively. After that, through a mapping layer $F_{M}$, the original visual features are projected into the text feature space to obtain aligned visual features. The text features and the aligned visual features are concatenated along their dimensions to form a cross-modal query vector $Q$:
\vspace{-0.3em}
\begin{equation}
	Q = \text{Concat}(F_L(q), F_M(F_V(i))) \in \mathbb{R}^{(l_q + l_i) \times d_L},
	\vspace{-0.3em}
\end{equation}
where $l_q$ denotes the sequence length of question $q$, and $l_i$ denotes the sequence length of the mapped visual features.

First Stage: Title-level multimodal coarse retrieval. Using the same text encoder $F_L$ as used for question encoding, we encode title $t \text{ (}t \in Title\text{)}$ to generate a title feature vector:
\vspace{-0.3em}
\begin{equation}
	T = F_L(t) \in \mathbb{R}^{l_t \times d_L},
	\vspace{-0.3em}
\end{equation}
where $l_t$ denotes the sequence length of title $t$. Based on the late interaction mechanism, finegrained relevance scores between the query vector $Q$ and each title $t$ are computed as:
\vspace{-1em}
\begin{equation}
	S(\overline{q}, t) = S((q,i), t) = \sum_{x=1}^{l_Q} \max_{y=1}^{l_T} {Q_x T_y^{\mathsf{T}}},
\end{equation}
where $l_Q = l_q + l_i$ denotes the sequence length of the query vectors, $Q_x$ denotes the $x$-th query embedding vector, and $T_y$ denotes the $y$-th embedding vector. The relevance scores for all titles are sorted in descending order, and the top $k_1$ titles corresponding to the highest scores are selected to form the candidate title set:
\vspace{-0.5em}
\begin{equation}
	T_{\text{top-}k_1} = \arg\max_{\substack{T' \subseteq Title \\ |T'|=k_1}} \left\{ S(\overline{q}, t) \mid t \in Title \right\}.
\end{equation}
Finally, we extract the union of all candidate passage subsets $P_{candidate}$ corresponding to the $M$ samples:
\begin{equation}
	P_{candidate} = \bigcup_{m=1}^{M} \left\{ \bigcup_{t \in T_{\text{top-}k_1}} P_t \right\}.
\end{equation}
At this point, the size of $P_{candidate}$ is much smaller than that of the full passage set $Passage$, yielding an efficient compression of the retrieval scope.

Second stage: Paragraph-level multimodal fine-grained retrieval. We take $P_{candidate}$ as the retrieval target, reuse the same relevance metric, and achieve precise recall of the target knowledge paragraphs. Specifically, we encode $p \text{ (}p \in P_{candidate}\text{)}$ using text encoder $F_L$:
\vspace{-0.3em}
\begin{equation}
	P = F_L(p) \in \mathbb{R}^{l_p \times d_L},
	\vspace{-0.3em}
\end{equation}
where $l_p$ denotes the sequence length of $p$. Applying the relevance scoring function defined in the first stage, the relevance score between the query vector $Q$ and paragraph feature $P$ is computed as:
\vspace{-0.3em}
\begin{equation}
	S(\overline{q}, p) = S((q,i), p) = \sum_{x=1}^{l_Q} \max_{y=1}^{l_P} {Q_x P_y^{\mathsf{T}}},
	\vspace{-0.3em}
\end{equation}
where $P_y$ denotes the $y$-th embedding vector. Finally, the relevance scores of the candidate paragraphs are sorted in descending order, and the knowledge paragraphs corresponding to the top-$k_2$ scores are selected to form the target knowledge fragment set for a single sample as:
\vspace{-0.3em}
\begin{equation}
	P_{\text{top-}k_2} = \arg\!\!\!\!\!\max_{\substack{P' \subseteq P_{candidate} \\ |P'|=k_2}} \left\{ S(\overline{q}, p) \mid p \in P_{candidate} \right\},
	\vspace{-0.3em}
\end{equation}
where $P_{\text{top-}k_2}$ represents the target knowledge paragraph set.

\subsection{Context Selection}
The Context Selection stage focuses on reducing evidence redundancy while preserving question-relevant semantics. LLMs are utilized to select visual knowledge and explicit knowledge from visual notes and top-$k_2$ explicit knowledge passages, respectively. 

Generally, for an image $I_{i}$ corresponding to question $Q_{i}$, along with the caption $C_{i}$, objects $O_{i}$, OCR string $S_{i}$, and explicit knowledge passages $P_{\text{top-}k_2}^i$ obtained from the CA stage, the selected visual knowledge is $VK_{i} = \mathcal{LLM} [\mathcal{P}(C_{i}, O_{i}, S_{i}, Q_{i})]$, the selected explicit knowledge is $EK_{i} = \mathcal{LLM} [\mathcal{P}(C_{i}, Q_{i}, P_{\text{top-}k_2}^i)]$, where $\mathcal{P}$ denotes the prompt template for this scene. Specifically, we feed both image captions and the question to retrieve implicit knowledge from LLMs to ensure alignment: $IK_{i} = \mathcal{LLM} [\mathcal{P}(C_{i}, Q_{i} )]$.

\subsection{Context Compression}
To bridge the gap between unstructured evidence and logical reasoning, we propose Relational Logic Distillation (RLD) in the Context Compression stage. The RLD is operationalized through a systematic logical decomposition procedure, which guides the LLM to transform textual evidence into explicit relational structures. RLD consists of three key steps: entity extraction, relation extraction, and relation refinement.

In the first step, we initialize the entity set $E_{Vi}$ using the detected objects $O_i$:
\begin{equation}
	E_{Vi} = \{e_{Vi}^{1}, ... , e_{Vi}^{k}\} \leftrightarrow O_{i} . 
\end{equation}
To avoid missing critical entities, we then use an LLM to separately extract the explicit and implicit knowledge entities from the sentence as follows: $E_{Ei} = \mathcal{LLM}[\mathcal{P}(VK_i,EK_i)], E_{Ii} = \mathcal{LLM}[\mathcal{P}(IK_i)]$. In consequence, the final explicit entity list can be expressed as $E_{i}^e = E_{Vi} \cup E_{Ei}$, the implicit entity list can be expressed as $E_{i}^i = E_{Vi} \cup E_{Ii}$.

In the second step, we extract the explicit and implicit relationships between entity pair $(e_{p}, e_{q})$ where $e_{p}$ $\ne$ $e_{q}$. The relation is denoted as triplets $(e_{p}, r, e_{q})$ where $r$ is the relation of entities $e_{p}$ and $e_{q}$. Specifically, for the $i$-th sample, its corresponding entity set $E_{i}^e$ and $E_{i}^i$, the initial explicit relation set $ER_{i}^{init} = \{..., (e_{p}, r_E, e_{q}) , ...\}$ ($e_{p}$, $e_{q} \in E_{i}^e$ and $e_{p}$ $\ne$ $e_{q}$) and the implicit relation set $IR_{i}^{init} = \{..., (e_{p}, r_I, e_{q}) , ...\}$ ($e_{p}$, $e_{q} \in E_{i}^i$ and $e_{p}$ $\ne$ $e_{q}$) are formulated as: 
\begin{equation}
	ER_{i}^{init} = \mathcal{LLM}[\mathcal{P}((VK_i,EK_i), E_{i}^e)],
\end{equation}
\begin{equation}
	IR_{i}^{init} = \mathcal{LLM}[\mathcal{P}(IK_i, E_{i}^i)].
\end{equation}
Therefore, the initial relations $R_i^{{init}} = ER_i^{{init}} \cup IR_i^{{init}}$.

In the final relation refinement step, we utilize an LLM to extract a list of key entities from the image caption and the question, and then filter the initial relation set by removing irrelevant relational triples. Specifically, for the visual knowledge $VK_i$ corresponding to question $Q_i$, the key entity list can be denoted as: $KE_i = \mathcal{LLM}[\mathcal{P}(VK_i, Q_i)]$. The final relations can be expressed as:  
\begin{equation}
	\begin{split}
		R_{i} = \bigl\{(&e_{p}, r, e_{q}) \bigm|
		(e_{p}, r, e_{q}) \in R_{i}^{{init}} \land [e_{p},  e_{q} \in KE_i  \\
		\lor\;&(\exists e_k \in KE_i,(e_p \overset{*}{\leftrightarrow} e_k) \;\lor\; (e_q \overset{*}{\leftrightarrow} e_k ))]\bigr\},
	\end{split}
\end{equation}
where $\overset{*}{\leftrightarrow}$ denotes that the two entities can reach each other through one or more relations. 

\subsection{Prediction}
In the final phase, the QA model predicts the answer by conditioning on the structured relation set $R_{i}$ and the question $Q_{i}$. By replacing unstructured context with explicit relational representations, the model can focus on reasoning over entity dependencies rather than on organizing raw information.

\section{Experiments}

\begin{table}[t]
	\centering
	\caption{Comparison with SOTA methods on the OK-VQA and A-OKVQA datasets. The best scores are highlighted in \textbf{bold}, and the second best scores are highlighted in \underline{underline}.}	
	\vspace{-0.5em}
	\begin{tabular}{l@{\hspace{-2pt}}c@{\hspace{-2pt}}ccc}
		\toprule
		\multirow{2}{*}{\centering Method} & \multicolumn{1}{c}{QA} & \multicolumn{1}{c}{\centering Shot}  & \multirow{2}{*}{\centering OK-VQA} & \multirow{2}{*}{\centering A-OKVQA} \\
		& Model & \centering Number &  &   \\
		\midrule
		\multicolumn{5}{c}{\textit{LLM-based Zero-shot Methods}} \\
		PICa \cite{yang2022empirical} & GPT-3$_{175B}$ & 0 &   17.1 & -  \\
		PNP-VQA \cite{tiong2022plug} & UnifiedQA$_{11B}$ & 0 &   35.9 & -  \\
		Img2LLM \cite{guo2023images} & OPT$_{175B}$ & 0 &  45.6 & 42.9  \\
		LAMOC \cite{du2023zero} & FLAN-T5-XXL$_{11B}$ & 0 &  40.3 & 37.9  \\
		RQ prompt \cite{lan2023improving} & GPT-3$_{175B}$ & 0 &  46.4 & 43.2  \\
		Emu \cite{emu} & Llama$_{13B}$ & 0  & 38.2 & -  \\
		ZVQAF \cite{Liu2024ZVQAFZV} & FLAN-T5$_{11B}$ & 0 &  40.5 & 37.1  \\
		KGenVQA \cite{cao2024knowledge} & UnifiedQA$_{11B}$ & 0 &  45.4 & 39.1  \\
		EF-VQA \cite{qiang2025enhancing} & GPT-3.5$_{175B}$ & 0 &  43.7 & 42.1  \\
		L2A \cite{xing2024learning} & GPT-3.5-turbo  & 0 &  46.2 & \underline{48.5}  \\
		PLLMKI \cite{pllmki} &  Llama$_{7B}$ & 0 &  28.4 & 24.8  \\
		DIETCOKE \cite{li2024diversify} & Gemma$_{7B}$ & 0  &  47.6 & 47.3  \\
		DIETCOKE \cite{li2024diversify} & Mistral$_{7B}$ & 0 &  \underline{49.2} & 47.5  \\
		\textbf{SCoRe (ours)} & Gemma$_{7B}$ & 0 &  49.4 & 50.3 \\
		\textbf{SCoRe (ours)} & Mistral$_{7B}$ & 0 &  \textbf{52.1} & \textbf{52.4} \\
		\midrule
		\multicolumn{5}{c}{\textit{LLM-based Few-shot Methods}} \\
		PICa \cite{yang2022empirical} & GPT-3$_{175B}$ & 1 &  40.8 & -  \\
		PICa \cite{yang2022empirical} & GPT-3$_{175B}$ & 4 &  45.4 & -  \\
		PICa \cite{yang2022empirical} & GPT-3$_{175B}$ & 16 &  48.0 & -  \\
		MM-R \cite{Khademi2023MMReasonerAM} & GPT-4-32k & 20 &  40.0 & -  \\
		VCTP \cite{Chen2024VisualCP} & OPT$_{66B}$ & 8 &  44.6 & 46.4  \\
		PCPA \cite{pcpa} & Llama$_{7B}$ & 16 &  53.6 & 52.3  \\
		PLLMKI \cite{pllmki} & Llama$_{7B}$ & 8 &  \underline{54.1} & \underline{52.9}  \\
		EF-VQA \cite{qiang2025enhancing} & GPT-3.5$_{175B}$ & 8 &  48.4 & 48.6  \\
		EF-VQA \cite{qiang2025enhancing} & GPT-3.5$_{175B}$ & 16 &  51.2 & 50.1  \\
		\textbf{SCoRe (ours)} & Mistral$_{7B}$ & 4 &  53.2 & 53.0  \\
		\textbf{SCoRe (ours)} & Mistral$_{7B}$ & 8 & 54.8 & 53.6 \\
		\textbf{SCoRe (ours)} & Mistral$_{7B}$ & 16 & \textbf{55.7} & \textbf{54.9} \\
		\bottomrule
	\end{tabular}
	\label{tab:zero_shot_vqa}
	\vspace{-1em}
\end{table}

\subsection{Experiment Settings}

\textbf{Datasets.}
We conduct experiments on the two popular KB-VQA datasets: Outside Knowledge VQA (OK-VQA) \cite{marino2019ok} and Augmented OK-VQA (A-OKVQA) \cite{schwenk2022okvqa}. Following EF-VQA \cite{qiang2025enhancing}, experimental evaluations are performed on the test set of OK-VQA, and the validation set of A-OKVQA.

\textbf{Implementation Details.}
DeepSeek-V3 is used as the context processor, while Mistral-7B and Gemma-7B are employed as the QA models. For DeepSeek-V3, we set the temperature to 0.7 and the presence penalty to 0.2. For Mistral-7B and Gemma-7B, we adopt a greedy decoding strategy. DeepSeek-V3 is accessed via API, whereas Mistral-7B and Gemma-7B are locally deployed on two RTX 3090 24 GB GPUs. We use VinVL \cite{zhang2021vinvl} for object detection, apply the captioning method from PNP-VQA \cite{tiong2022plug} to generate question-guided captions, and extract image text with the Google OCR API. In addition, we adopt the pre-trained PREFLMR \cite{preflmr} as the backbone for multimodal retrieval.

\textbf{Baselines.}
The methods we compare are categorized as follows: (1) LLM-based zero-shot methods: PNP-VQA \cite{tiong2022plug}, Img2LLM \cite{guo2023images}, LAMOC \cite{du2023zero}, RQprompt \cite{lan2023improving}, Emu \cite{emu}, ZVQAF \cite{Liu2024ZVQAFZV}, KGenVQA \cite{cao2024knowledge}, L2A \cite{xing2024learning}, and DIETCOKE \cite{li2024diversify}. (2) LLM-based few-shot methods: PICa \cite{yang2022empirical}, MM-R \cite{Khademi2023MMReasonerAM}, VCTP \cite{Chen2024VisualCP}, PCPA \cite{pcpa}, PLLMKI \cite{pllmki}, and EF-VQA \cite{qiang2025enhancing}. We follow official evaluation protocols and report VQA scores for the datasets. The accuracy metric is defined as $VQA\text{ }score= min(1, T(a) / 3)$, where $T(a)$ is the number of times the predicted answer appears in groundtruth. Notably, we choose the direct answer evaluation of A-OKVQA dataset.

\textbf{Zero-shot Settings.} We follow Img2LLM \cite{guo2023images} to generate QA pairs from image captions as demonstrations for zero-shot scenario. Throughout the entire process, the answer and any other samples remain untouched.

\textbf{Few-shot Settings.} We select the top-$n$ training samples as exemplars based on the average cosine similarity between CLIP-based image and question embeddings for reasoning.

\subsection{Performance of the Approaches}
The experimental results summarized in Table \ref*{tab:zero_shot_vqa} show that SCoRe consistently outperforms all baselines across both zero-shot and few-shot settings on OK-VQA and A-OKVQA. In the zero-shot scenario, SCoRe (Mistral-7B) achieves 52.1\% and 52.4\%, surpassing the previous best method, DIETCOKE, by a substantial margin. For the few-shot settings, SCoRe demonstrates steady performance gains as the shot count increases. At 16-shot, it reaches peak accuracies of 55.7\% and 54.9\%, establishing a new state-of-the-art. These consistent improvements across multiple benchmarks and settings underscore that transforming heterogeneous context into structured relational representations is more effective for knowledge-intensive reasoning than relying on unstructured text.

We also compare the time cost of SCoRe and DIETCOKE on Mistral-7B. SCoRe takes approximately 17.5 seconds per sample, whereas DIETCOKE requires 29.5 seconds. For visual question answering tasks, such time costs are acceptable while achieving improved accuracy.
\begin{figure}[t]
	\centering
	\includegraphics[width=0.75\columnwidth]{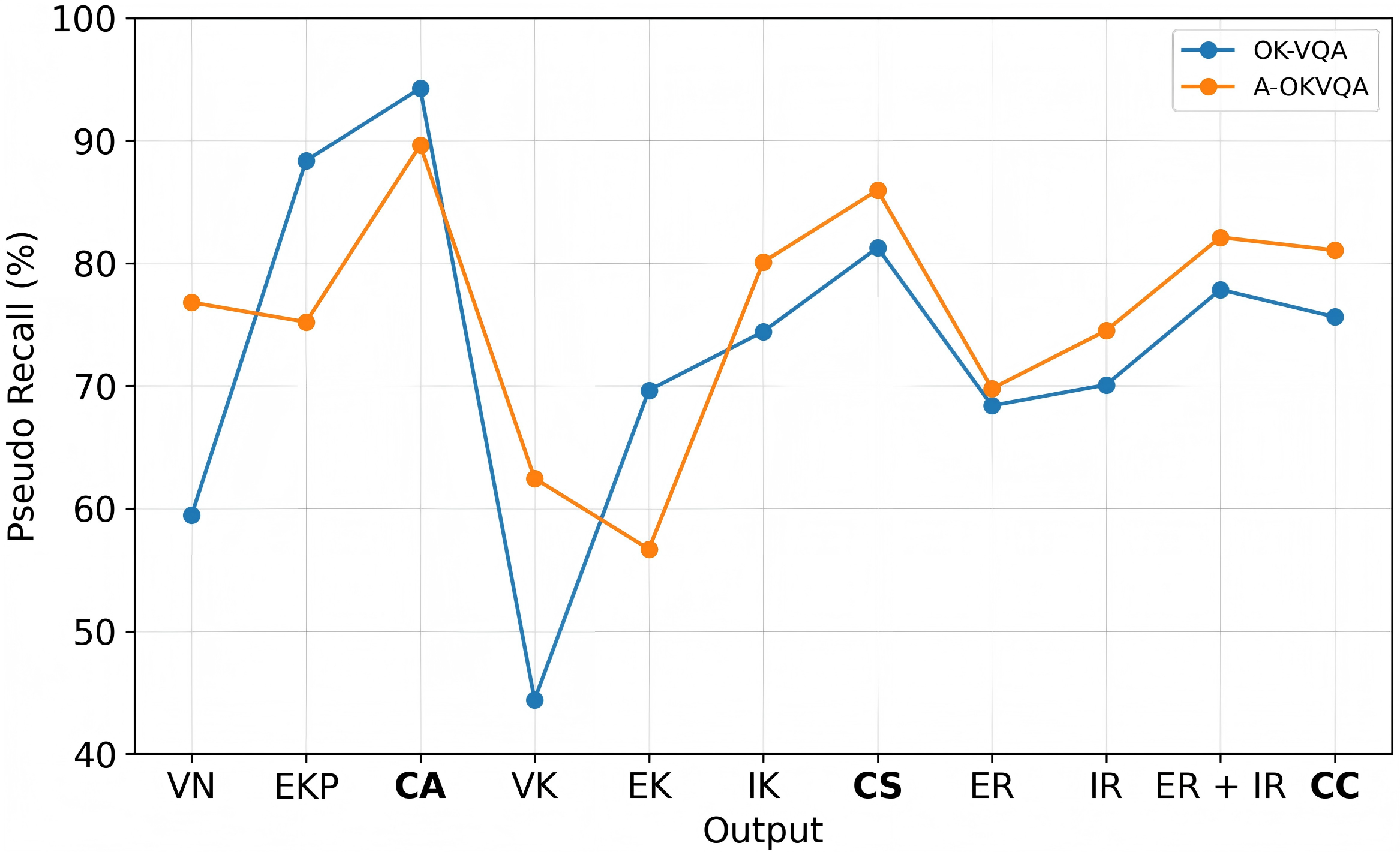}
	\vspace{-0.5em}
	\caption{The Pseudo Recall of intermediate outputs.}
	\label{fig:pr}
	\vspace{-0.5em}
\end{figure}

\begin{table}[!t]
	\centering
	\caption{Ablation study results of key modules on OK-VQA dataset.}
	\vspace{-0.5em}
	\begin{tabular}{ccccc}
		\toprule
		VN & EKP & CS & CC & OK-VQA \\
		\midrule
		\checkmark & & & & 47.5 \\
		\checkmark & \checkmark & & & 49.0 \\
		\checkmark & \checkmark & \checkmark & & 51.2 \\
		\checkmark & \checkmark & \checkmark & \checkmark & \textbf{52.1} \\
		\bottomrule
	\end{tabular}
	\label{tab:ablation_module}
	\vspace{-1em}
\end{table}

\subsection{Ablation Studies}

\textbf{Effectiveness of Proposed Modules.} We conduct an ablation study on the OK-VQA dataset to validate the effectiveness of each module in SCoRe. As shown in Table \ref{tab:ablation_module}, we gradually incorporate each component to evaluate the performance improvement. Using only visual notes (VN) yields 47.5\%. Adding explicit knowledge passages (EKP) further improves the result to 49.0\%. With the introduction of the CS module, the score increases to 51.2\%. Finally, equipping with the CC module achieves the best performance of 52.1\%. These results demonstrate that each proposed module contributes to the final performance, and the full model with all components delivers the optimal reasoning ability.

\textbf{Effects of Parameters.} We use the Pseudo Recall (PR) to evaluate the impact of $k_1$ and $k_2$ on the retrieval results, where the PR is defined as the proportion of knowledge passages that contain the correct answer among all the retrieved passages. Experimental results on A-OKVQA validation set (1145 samples) are presented in Table \ref{tab:pr} and Table \ref{tab:num}. Table \ref{tab:pr} shows that performance is already strong when $k_1$=1 and improves steadily as the number of titles increases. Beyond $k_1$=50 the gain tapers off, and PR@20 is almost identical for $k_1$=100 and $k_1$=200. Balancing efficiency and accuracy, we set $k_1$=200, $k_2$=10. Table \ref{tab:num} lists the counts of candidate passages for different $k_1$ values. We observe that our method compresses the retrieval counts by tens to hundreds of times while still delivering promising results.

\begin{table}[!t]
	\centering
	\caption{The Pseudo Recall across different $k_1$ and $k_2$.}
	\vspace{-0.5em}
	\begin{tabular}{c@{\hspace{0.7em}}c@{\hspace{0.7em}}c@{\hspace{0.7em}}c@{\hspace{0.7em}}c@{\hspace{0.7em}}c@{\hspace{0.7em}}c}
		\toprule
		top-$k$ & $k_1$=1& $k_1$=10 & $k_1$=20 & $k_1$=50 & $k_1$=100 & $k_1$=200 \\
		\midrule
		$k_2$=5 (PR@5) & 55.02 & 56.41& 58.77& 63.58& 63.58& 64.62\\
		$k_2$=10 (PR@10) & 66.72 & 68.64& 68.90& 73.71& 74.14& 75.19\\
		$k_2$=20 (PR@20) & 76.76 & 78.07& 77.99& 81.74& 82.88& 82.88 \\
		\bottomrule
	\end{tabular}
	\label{tab:pr}
	\vspace{-1em}
\end{table}

\begin{table}[!t]
	\centering
	\caption{The number of candidate passages for different $k_1$.}
	\vspace{-0.5em}
	\begin{tabular}{l@{\hspace{0.5em}}c@{\hspace{0.5em}}c@{\hspace{0.5em}}c@{\hspace{0.5em}}c@{\hspace{0.5em}}c@{\hspace{0.5em}}c@{\hspace{0.5em}}c}
		\toprule
		& ALL & $k_1$=1& $k_1$=10 & $k_1$=20 & $k_1$=50 & $k_1$=100 & $k_1$=200 \\
		\midrule
		Num & 21,015,324 & 8,746 & 72,969& 135,835& 294,726 & 560,324& 988,434\\
		\bottomrule
	\end{tabular}
	\label{tab:num}
	\vspace{-1em}
\end{table} 

\textbf{Information Loss of Intermediate Output.} As SCoRe adopts a pipeline-based architecture, each stage of content generation or extraction inevitably alters and partially loses information. Fig. \ref{fig:pr} shows the PR of intermediate outputs, including Visual Notes (VN), Explicit Knowledge Passages (EKP), CA (VN + EKP), Visual Knowledge (VK), Explicit Knowledge (EK), Implicit Knowledge (IK), CS (VK + EK + IK), Explicit Relations (ER), Implicit Relations (IR), ER + IR, and CC (refined ER + IR). As can be seen from the figure, the PR at the initial CA stage is approximately 10 percentage points higher than that at the final CC stage on two datasets. However, the actual performance of the CA is lower than the CC. This is because CA contains substantial redundant and noisy information, which impairs the model's judgment. In contrast, the structured information in the CC stage enables the model to better understand the relations between entities, thereby deriving accurate answers. 

\subsection{Case study} 
\begin{figure*}[!t]
	\centering
	\includegraphics[width=\textwidth]{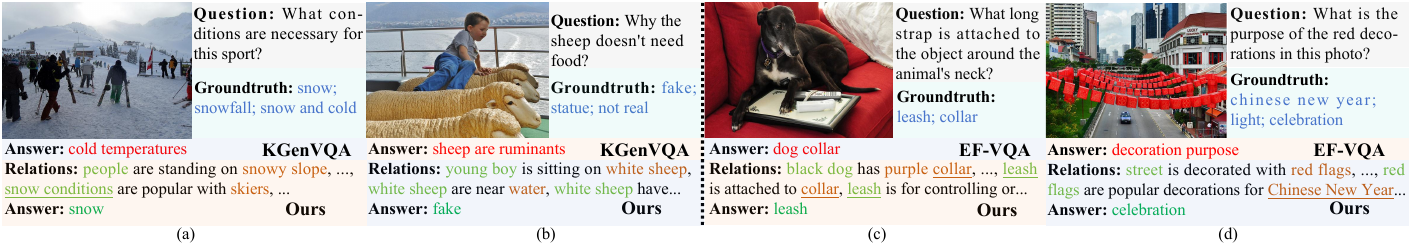}
	\vspace{-1.6em}
	\caption{The qualitative comparison of KGenVQA with zero-shot settings (a-b) and EF-VQA with few-shot settings (c-d).}
	\label{fig:examples}
	\vspace{-1.0em}
\end{figure*}

As shown in Fig. \ref{fig:examples}, we select some representative cases from OK-VQA to compare SCoRe with KGenVQA (a-b, zero-shot) and EF-VQA (c-d, few-shot). Baselines relying on unstructured implicit knowledge often suffer from noisy or incorrect reasoning. In contrast, SCoRe extracts structured relation triplets via context reasoning, integrating visual, explicit and implicit knowledge. SCoRe outputs accurate answers (e.g., ``snow", ``celebration") by capturing key info and logical connections, verifying its superiority in KB-VQA.

\begin{figure}[!t]
	\centering
	\includegraphics[width=\columnwidth]{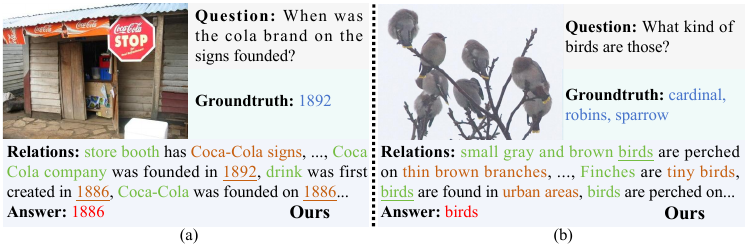}
	\vspace{-1.6em}
	\caption{Qualitative case study of SCoRe failure modes.}
	\label{fig:wrong_cases}
	\vspace{-1.0em}
\end{figure}

Despite its effectiveness, SCoRe occasionally encounters limitations, as illustrated in the failure cases. In Fig. \ref{fig:wrong_cases} (a), although the explicit knowledge correctly retrieves ``1892", the model outputs ``1886" due to a conflict between explicit and implicit relations. The LLM’s internal knowledge regarding the formula's invention (1886) overrides the specific brand founding date (1892) present in the retrieved context, leading to an incorrect reasoning priority. In Fig. \ref{fig:wrong_cases} (b), the model provides a generic answer ``birds" instead of specific species like ``cardinal". This failure is attributed to insufficient finegrained visual knowledge, while the SCoRe framework identifies the presence of birds and their basic colors, it lacks the specialized ornithological features (e.g., subtle beak shapes or plumage patterns) necessary for precise taxonomic classification.

\section{Conclusion}
In this paper, we propose an LLM-based Structured Context Reasoning (SCoRe) framework, which extracts structured relationships for knowledge-based visual question answering. SCoRe addresses two pivotal challenges by structured context reasoning to refine the original information: 1) Excessive inputs cognitively overload LLMs, and 2) Difficulty in reasoning over unstructured input. In addition, SCoRe introduces a two-stage progressive multimodal retrieval strategy that dramatically improves retrieval efficiency while maintaining high accuracy. Extensive experiments demonstrate that SCoRe outperforms state-of-the-art methods.

\section*{Acknowledgments}
This work is supported by the Guangdong Basic and Applied Basic Research Foundation (No.2024A1515010237),  and a grant from CMA Communication and Outreach Centre (China Meteorological News Press) Meteorological Converged Media Technology Innovation and Application Open Laboratory (No. QXRM-ZD-2508).

\bibliographystyle{IEEEtran}
\bibliography{IEEEexample}

@inproceedings{lin2022retrieval,
	title={Retrieval Augmented Visual Question Answering with Outside Knowledge},
	author={Lin, Weizhe and Byrne, Bill},
	booktitle={EMNLP},
	year={2022}
}

@inproceedings{yang2022empirical,
	title={An empirical study of gpt-3 for few-shot knowledge-based vqa},
	author={Yang, Zhengyuan and Gan, Zhe and Wang, Jianfeng and Hu, Xiaowei and Lu, Yumao and Liu, Zicheng and Wang, Lijuan},
	booktitle={AAAI},
	year={2022}
}

@inproceedings{guo2023images,
	title={From images to textual prompts: Zero-shot visual question answering with frozen large language models},
	author={Guo, Jiaxian and Li, Junnan and Li, Dongxu and Tiong, Anthony Meng Huat and Li, Boyang and Tao, Dacheng and Hoi, Steven},
	booktitle={CVPR},
	year={2023}
}

@article{qiang2025enhancing,
	title={Enhancing few-shot KB-VQA with panoramic image captions guided by Large Language Models},
	author={Qiang, Pengpeng and Tan, Hongye and Li, Xiaoli},
	journal={Neurocomputing},
	year={2025},
	publisher={Elsevier}
}

@inproceedings{li2024diversify,
	title={Diversify, Rationalize, and Combine: Ensembling Multiple QA Strategies for Zero-shot Knowledge-based VQA},
	author={Li, Miaoyu and Li, Haoxin and Du, Zilin and Li, Boyang},
	booktitle={EMNLP},
	year={2024}
}

@inproceedings{cao2024knowledge,
	title={Knowledge Generation for Zero-shot Knowledge-based VQA},
	author={Cao, Rui and Jiang, Jing},
	booktitle={EACL},
	year={2024}
}

@inproceedings{zhang2021vinvl,
	title={Vinvl: Revisiting visual representations in vision-language models},
	author={Zhang, Pengchuan and Li, Xiujun and Hu, Xiaowei and Yang, Jianwei and Zhang, Lei and Wang, Lijuan and others},
	booktitle={CVPR},
	year={2021}
}

@inproceedings{tiong2022plug,
	title={Plug-and-Play VQA: Zero-shot VQA by Conjoining Large Pretrained Models with Zero Training},
	author={Tiong, Anthony Meng Huat and Li, Junnan and Li, Boyang and Savarese, Silvio and Hoi, Steven CH},
	booktitle={EMNLP},
	year={2022}
}

@inproceedings{liu2024era,
	title={ERA-CoT: Improving Chain-of-Thought through Entity Relationship Analysis},
	author={Liu, Yanming and Peng, Xinyue and Du, Tianyu and Yin, Jianwei and Liu, Weihao and Zhang, Xuhong},
	booktitle={ACL},
	year={2024}
}

@inproceedings{marino2019ok,
	title={Ok-vqa: A visual question answering benchmark requiring external knowledge},
	author={Marino, Kenneth and Rastegari, Mohammad and Farhadi, Ali and Mottaghi, Roozbeh},
	booktitle={CVPR},
	year={2019}
}

@inproceedings{schwenk2022okvqa,
	title={A-okvqa: A benchmark for visual question answering using world knowledge},
	author={Schwenk, Dustin and Khandelwal, Apoorv and Clark, Christopher and Marino, Kenneth and Mottaghi, Roozbeh},
	booktitle={ECCV},
	year={2022}
}

@inproceedings{du2023zero,
	title={Zero-shot Visual Question Answering with Language Model Feedback},
	author={Du, Yifan and Li, Junyi and Tang, Tianyi and Zhao, Wayne Xin and Wen, Ji-Rong},
	booktitle={ACL},
	year={2023}
}

@article{Liu2024ZVQAFZV,
	title={ZVQAF: Zero-shot visual question answering with feedback from large language models},
	author={Cheng Liu and Chao Wang and Yan Peng and Zhixu Li},
	journal={Neurocomputing},
	year={2024},
}

@inproceedings{Khademi2023MMReasonerAM,
	title={MM-Reasoner: A Multi-Modal Knowledge-Aware Framework for Knowledge-Based Visual Question Answering},
	author={Mahmoud Khademi and Ziyi Yang and Felipe Vieira Frujeri and Chenguang Zhu},
	booktitle={EMNLP},
	year={2023}
}

@inproceedings{Chen2024VisualCP,
	title={Visual Chain-of-Thought Prompting for Knowledge-Based Visual Reasoning},
	author={Zhenfang Chen and Qinhong Zhou and Yikang Shen and Yining Hong and Zhiqing Sun and Dan Gutfreund and Chuang Gan},
	booktitle={AAAI},
	year={2024}
}

@inproceedings{lan2023improving,
	title={Improving zero-shot visual question answering via large language models with reasoning question prompts},
	author={Lan, Yunshi and Li, Xiang and Liu, Xin and Li, Yang and Qin, Wei and Qian, Weining},
	booktitle={ACM MM},
	year={2023}
}

@inproceedings{xing2024learning,
	title={Learning to Ask Denotative and Connotative Questions for Knowledge-based VQA},
	author={Xing, Xiaoying and Xiong, Peixi and Fan, Lei and Li, Yunxuan and Wu, Ying},
	booktitle={EMNLP},
	year={2024}
}

@inproceedings{wu2022multi,
	title={Multi-modal answer validation for knowledge-based vqa},
	author={Wu, Jialin and Lu, Jiasen and Sabharwal, Ashish and Mottaghi, Roozbeh},
	booktitle={AAAI},
	year={2022}
}

@inproceedings{shao2023prompting,
	title={Prompting large language models with answer heuristics for knowledge-based visual question answering},
	author={Shao, Zhenwei and Yu, Zhou and Wang, Meng and Yu, Jun},
	booktitle={CVPR},
	year={2023}
}

@inproceedings{preflmr,
	title = "{P}re{FLMR}: Scaling Up Fine-Grained Late-Interaction Multi-modal Retrievers",
	author = "Lin, Weizhe  and
	Mei, Jingbiao  and
	Chen, Jinghong  and
	Byrne, Bill",
	booktitle = "ACL",
	year = "2024",
}

@inproceedings{emu,
	author = {Sun, Quan and Yu, Qiying and Cui, Yufeng and Zhang, Fan and Zhang, Xiaosong and Wang, Yueze and Gao, Hongcheng and Liu, Jingjing and Huang, Tiejun and others},
	booktitle = {ICLR},
	title = {Emu: Generative Pretraining in Multimodality},
	year = {2024}
}

@article{pcpa,
	title = {Prompting large language model with context and pre-answer for knowledge-based VQA},
	journal = {Pattern Recognition},
	year = {2024},
	author = {Zhongjian Hu and Peng Yang and Yuanshuang Jiang and Zijian Bai},
}

@article{pllmki,
	author={Hu, Zhongjian and Yang, Peng and Liu, Fengyuan and Meng, Yuan and Liu, Xingyu},
	journal={Big Data Mining and Analytics}, 
	title={Prompting Large Language Models with Knowledge-Injection for Knowledge-Based Visual Question Answering}, 
	year={2024},
}

@inproceedings{osllm,
	author={Zhou, Jie and Shan, Yongxue and Wu, Meihan and Hu, Fei and Zheng, Li and Wang, Xiaodong},
	booktitle={ICME}, 
	title={OSLLM: A Retrieve-Reason-Refine Framework for Multi-Domain Relation Extraction with Large Language Models}, 
	year={2025},}

@inproceedings{mumrc,
	author={Chen, Qiang and Zhang, Dong and Li, Shoushan and Zhou, Guodong},
	booktitle={ICME}, 
	title={A Unified MRC Framework with Multi-Query for Multi-modal Relation Triplets Extraction}, 
	year={2023}}

@inproceedings{tevla,
	author={Chen, Junlin and Guo, Qiushan and Cheung, Ka Chun and Liang, Mingrui and Chen, Dezhi},
	booktitle={ICME}, 
	title={TEVLA: Text-oriented Enhancement for Vision-Language Alignment in Relation Extraction}, 
	year={2025}}

@inproceedings{qa4re,
	title = {Aligning Instruction Tasks Unlocks Large Language Models as Zero-Shot Relation Extractors},
	author = {Zhang, Kai  and
	Jimenez Gutierrez, Bernal  and
	Su, Yu},
	booktitle = {ACL},
	year = {2023}
}

\end{document}